\documentclass[sigconf]{acmart}

\usepackage{booktabs} 
\usepackage{multirow}
\usepackage{indentfirst} 
\usepackage{flushend}
\usepackage{enumitem}
\usepackage[bf, medium]{titlesec}
\usepackage{url}

\DeclareMathOperator*{\argmax}{arg\,max}

\usepackage{subfigure}

\titlespacing\section{0pt}{7pt minus 2pt}{0 pt plus 2pt}
\titlespacing\subsection{0pt}{7pt minus 2pt}{0pt plus 2pt} 
\titlespacing\subsubsection{0pt}{7pt minus 0pt}{0pt plus 2pt}

\setcopyright{none}

\begin{document}
\title{Video Captioning with Guidance of Multimodal Latent Topics}

\newcommand{\superscript}[1]{\ensuremath{^{\textrm{#1}}}}

\author{Shizhe Chen}
\affiliation{%
  \institution{Renmin University of China}
}
\email{cszhe1@ruc.edu.cn}

\author{Jia Chen}
\affiliation{%
  \institution{Carnegie Mellon University}
}
\email{jiac@cs.cmu.edu}

\author{Qin Jin}
\authornote{corresponding author.}
\affiliation{%
  \institution{Renmin University of China}
}
\email{qjin@ruc.edu.cn}

\author{Alexander Hauptmann}
\affiliation{%
  \institution{Carnegie Mellon University}
}
\email{alex@cs.cmu.edu}

\begin{abstract}
\noindent
The topic diversity of open-domain videos leads to various vocabularies and linguistic expressions in describing video contents, and therefore, makes the video captioning task even more challenging. 
In this paper, we propose an unified caption framework, M\&M TGM, which mines multimodal topics in unsupervised fashion from data and guides the caption decoder with these topics. 
Compared to pre-defined topics, the mined multimodal topics are more semantically and visually coherent and can reflect the topic distribution of videos better. 
We formulate the topic-aware caption generation as a multi-task learning problem, in which we add a parallel task, topic prediction, in addition to the caption task.
For the topic prediction task, we use the mined topics as the teacher to train a student topic prediction model, which learns to predict the latent topics from multimodal contents of videos. The topic prediction provides intermediate supervision to the learning process.
As for the caption task, we propose a novel topic-aware decoder to generate more accurate and detailed video descriptions with the guidance from latent topics.
The entire learning procedure is end-to-end and it optimizes both tasks simultaneously.
The results from extensive experiments conducted on the MSR-VTT and Youtube2Text datasets demonstrate the effectiveness of our proposed model.
M\&M TGM not only outperforms prior state-of-the-art methods on multiple evaluation metrics and on both benchmark datasets,
but also achieves better generalization ability.
\end{abstract}

%
%
%



\maketitle

\section{Introduction}
Videos have become increasingly popular on the Internet, 
for example, hundreds of hours of video contents are uploaded on YouTube every minute. 
It is impossible to watch these overwhelming amounts of videos, therefore, automatic techniques to search and analyze video contents are highly desired. 
Generating natural language descriptions for video contents (a.k.a. video captioning) is one of such important techniques for this challenge. 
It can benefit a wide range of applications such as assisting the visually impaired people and improving the quality of online video retrieval.

Although there have been significant breakthroughs recently in image captioning \cite{DBLP:conf/cvpr/VinyalsTBE15,xu2015show,you2016image,Gan2016Semantic,yang2016review,rennie2016self},
video captioning remains very challenging due to the diversity and complexity of video contents. 
The open-domain videos contain a broad range of topics, such as sports, music, food and so on,
which results in very different vocabularies and expression styles to describe video contents across topics.
For example, sports terms frequently occur in the sports topic while the typical words in the food topic are about cooking and ingredients.
The key contents in descriptions for the sports topic are often the action verbs, while they are object nouns for the food topic and descriptive adjectives for the people topic.
Thus, being aware of the topic information can dramatically narrow down general sentence distributions and enable the caption model to focus on the discriminative video contents under the topic, such as the detailed actions in the sports topic.

However, most of the existing caption models ignore the topic information and mainly try to maximize the overall likelihood for videos in all topics,
which have a tendency to seek the most common mode in training sentences as indicated by Lee et al. \cite{lee2016stochastic}.
Such models not only are prone to generate plain descriptions without details, but also unable to distinguish confusing concepts in a topic.
Although a few works \cite{DBLP:conf/mm/JinCCXH16,cvpr17densevideocaption,dong2017improving} have exploited the topic information for video description generation,
there are still three main challenges in employing the topic information in video captioning. 
First, there are no direct topic representations in common video caption datasets, so how can we construct and predict the latent topics of videos?
Second, what is an effective and efficient way to employ the latent topics to guide the caption model for generating topic-oriented descriptions?
And third, how to train the topic-guided caption model to achieve optimal captioning performance?

In this paper, we bring three innovations to deal with the above challenges.
First, we propose a multimodal topic mining approach to discover latent video topics from video-sentence pairs.
The multimodal latent topics are more semantically coherent and visually consistent than the expert-defined or textual latent topics in previous work.
We then use the teacher-student learning strategy to predict latent topics with multimodal video features.
Second, in order to exploit the topic effectively, we propose a novel topic-aware language decoder, which implicitly functions as an ensemble of topic-specific decoders for each topic, but is computationally more efficient and requires less training data.
The predicted video topic automatically modifies weights in the decoder, enabling to generate topic-oriented video descriptions.
Third, to optimize the topic-guided caption performance, we develop a multi-task learning architecture to jointly train the caption system with the topic prediction loss and sentence generation loss in an end-to-end manner.
It can strengthen the coupling of the two parts to generate better latent topics and sentence descriptions simultaneously. 
The overall model is called \textbf{M\&M TGM} (Topic-Guided Model with Multimodal latent topic guidance and Multi-task learning).

The main contributions of this paper are as follows:
\begin{itemize}[leftmargin=10pt]
\item We present an unified framework M\&M TGM for video captioning, which can automatically discover and predict the underlying video topics and exploit the topic guidance to generate better topic-oriented video descriptions.
The topic prediction and topic-aware sentence generation can be trained jointly with multi-task learning end-to-end.

\item The proposed M\&M TGM requires no additional labelling for video topics and can exploit the limited video caption data effectively.
For the latent topic generation, we propose unsupervised multimodal topic mining for topic discovery and teacher-student learning for topic prediction.
For the topic-oriented sentence generation, the topic-aware decoder shares data and parameters among different topics which is efficient in both computation and data.

\item Our proposed model is evaluated on two benchmark video caption datasets: MSR-VTT \cite{xu2016msr} and Youtube2Text \cite{DBLP:conf/iccv/GuadarramaKMVMDS13}. 
Both quantitative and qualitative analysis show superior performances of using the multimodal latent topic guidance and multi-task learning strategy.
The M\&M TGM not only outperforms prior state-of-the-art methods, but also has better generalization ability.
\end{itemize}

The rest of the paper is organized as follows:
Section 2 introduces the related work.
Section 3 presents the problem formulation.
The solution for M\&M TGM is described in Section 4.
Section 5 presents experimental results and analysis.
Section 6 concludes the paper. 
\section{Related Work}

\textbf{Image Captioning} has attracted growing interests recently.
Early works are mainly based on rule systems \cite{kulkarni2011baby,DBLP:conf/icml/LebretPC15} and suffer from generating flexible and accurate descriptions.
So more researches have been focusing on the encoder-decoder framework \cite{bahdanau2014neural,sutskever2014sequence} for caption generation.
The deep convolutional neural networks (CNNs) \cite{Szegedy2016Inception} function as the encoder to transform image contents into dense vectors.
Then the decoder, typically the Long-short Term Memory \cite{hochreiter1997long} (LSTM), is utilized to generate sequential words conditioned on image features \cite{DBLP:conf/cvpr/VinyalsTBE15}.
Xu et al. \cite{xu2015show} propose the spatial attention mechanism based on the basic encoder-decoder.
You et al. \cite{you2016image} and Gan et al. \cite{Gan2016Semantic} propose the semantic attention model and semantic compositional networks (SCN) respectively to exploit detected concepts from the image.
Our topic-aware language decoder is inspired by SCN but with different aims of producing topic-oriented video descriptions, which is more suitable for video captioning task as shown in Section~\ref{sec:state-of-art-cmpr}.

\textbf{Video Captioning} is more challenging compared with image captioning due to the temporal dynamics, multiple modalities, and more diverse contents in videos.
Yao et al. \cite{DBLP:conf/iccv/YaoTCBPLC15} propose the temporal attention mechanism,
and Pan et al. \cite{pan2015hierarchical} utilize the hierarchical LSTM encoder to explore the temporal structures.
Most previous works only focus on the visual modality \cite{DBLP:conf/cvpr/PanMYLR16,Venugopalan2014Translating},
but recently Jin et al. \cite{jin2016video,DBLP:conf/mm/JinCCXH16} and Ramanishka et al. \cite{ramanishka2016multimodal} have shown improvement of multimodal fusion for video captioning.
To promote the video captioning research for open-domain videos, several large-scale videos with various topic are collected such as MSR-VTT \cite{xu2016msr} and TGIF \cite{li2016tgif}.
Jin et al. \cite{DBLP:conf/mm/JinCCXH16} encode expert-defined topics with multimodal features, which results in their winning of the MSR-VTT challenge \cite{VTTGC2016}.
Dong et al. \cite{dong2017improving} utilize the textual mined topics to learn interpretable features.
Shen et al. \cite{cvpr17densevideocaption} train separate language decoders for each expert-defined topic.
Chen et al. \cite{DBLP:conf/mir/ChenCJ17} explore the guidance from textual minded topics to generate topic-aware sentences.
In our work, we address the topic diversity challenge in open-domain videos and propose the novel M\&M TGM model to jointly generate the latent video topics and topic-oriented video descriptions with multimodal features in a multi-task framework.

\textbf{Latent Topic Mining} has been a long-standing problem. 
The latent Dirichlet allocation (LDA) proposed by Blei \cite{blei2003latent} is one of the most classic models to automatically inference the latent topics for textual documents.
Doersch et al. \cite{doersch2012makes} utilize a discriminative clustering approach to discover the visual topics.
For multimodal data, Li et al. \cite{li2016event} apply the association rule mining algorithm on image-caption pairs to discover the multimodal topics.
In our work, we take the multimodal topic mining perspective on video-sentence pairs by a weighted multimodal clustering method to obtain semantically coherent and visually consistent latent topics.

\section{Problem Formulation}
\label{sec:problem_form}

Our goal is to leverage the underlying video topic information to make the model more proficient in using vocabularies and expressions within the topic when describing a video.

Suppose we have a video $\mathcal{V}$ along with a set of $N_{d}$ textual descriptions $\mathrm{Y} = \{\mathrm{y}_{1}, \mathrm{y}_{2}, ..., \mathrm{y}_{N_{d}}\}$.
Each $\mathrm{y} \in \mathrm{Y}$ is a sentence with $N_{s}$ words, denoted as $\mathrm{y} = \{w_{1}, w_{2}, ..., w_{N_{s}}\}$, where $w$ is a word from vocabulary $\mathrm{W}$.
We encode the video content $\mathcal{V}$ into a fixed-dimensional video representation $\mathrm{x}$.
Conditioned on the multimodal video representation $\mathrm{x}$, a caption decoder aims to generate the sentence $\mathrm{y}$ with probability
\begin{equation}
\label{eq:language}
\mathrm{Pr(y|x)} = \prod_{t=1}^{N_s} \mathrm{Pr}(w_t|\mathrm{x}, w_{<t})
\end{equation}
where $w_{<t}$ denotes the sequential words before the time step $t$.
We abbreviate this as \emph{conditional sentence probability}. 
We use RNN as decoder and parameterize the probability $\mathrm{Pr}(w_t|\mathrm{x}, w_{<t})$ by the recurrent units shared across time steps, which can be expressed as:
\begin{equation}
\label{eqn:rnn_decoder}
\left [
\begin{array}{c}
\mathrm{Pr}(w_t|\mathrm{x}, w_{<t}; \Theta_d)\\
h_t
\end{array}
\right ]
= \psi_{d}(h_{t-1}, w_{t-1}; \Theta_d)
\end{equation}
where $\psi_{d}$ is the recurrent unit function, $\Theta_{d}$ are the parameters of the recurrent unit and $h_t$ is the hidden state of the recurrent unit at time step $t$. 
We define $w_0$ as the start token and $h_0$ as the initialized hidden state.
For notation simplicity, we abbreviate eq (\ref{eqn:rnn_decoder}) to $\mathrm{Pr}(w_t|\mathrm{x}, w_{<t}; \Theta_d) = \psi_{d}(h_{t-1}, w_{t-1}; \Theta_d)$. Substituting this abbreviation to eq~\eqref{eq:language}, we get:
\begin{equation}
\label{eq:sentence_parameterize}
\mathrm{Pr(y|x}; \Theta_d) = \prod_{t=1}^{N_s} \psi_{d}(h_{t-1}, w_{t-1}; \Theta_d)
\end{equation}

As mentioned earlier, videos in different topics have quite different vocabularies and expression styles in their video descriptions,
and different topics can coexist in one video.
Based on this observation, we introduce the latent topic variables $\{\mathrm{z}_i\}_{i = 1 , \dots, K}$, where $K$ is the number of topics.
Then the conditional sentence probability, $\mathrm{Pr}(\mathrm{y} | \mathrm{x})$ is actually the marginal probability of the joint sentence and topic distribution conditioned on video content:
\small
\begin{equation}
\label{eq:topic_aware}
\begin{aligned}
&\mathrm{Pr}(\mathrm{y}|\mathrm{x}) = \sum_{\mathrm{z}_1, \dots, \mathrm{z}_K} \mathrm{Pr}(\mathrm{y}, \mathrm{z}_1, \dots, \mathrm{z}_K|\mathrm{x}) \\
&= \sum_{\mathrm{z}_1, \dots, \mathrm{z}_K} \underbrace{\mathrm{Pr}(\mathrm{y}|\mathrm{z}_1, \dots, \mathrm{z}_K,\mathrm{x})}_{\substack{\text{topic-aware} \\ \text{sentence distribution}}} \underbrace{\mathrm{Pr}(\mathrm{z}_1, \dots, \mathrm{z}_K|\mathrm{x})}_{\substack{\text{latent topic} \\ \text{distribution}}} \\
\end{aligned}
\end{equation} 
\normalsize
In the second step of eq (\ref{eq:topic_aware}), we factorize the joint distribution to latent topic distribution and topic-aware sentence distribution. 

We come to parameterize the probability in eq (\ref{eq:topic_aware}).
For latent topic distribution, we parameterize it by \emph{topic predictor} $\psi_{z}$ as follows:
\small
\begin{equation}
\mathrm{Pr}(\mathrm{z}_1, \dots, \mathrm{z}_K|\mathrm{x}) = \psi_{z}(\mathrm{x}; \Theta_z)
\end{equation}
\normalsize
where $\Theta_{z}$ are parameters in the topic predictor. 
To parameterize the topic-aware sentence distribution by RNN, we need to introduce the topic-related parameter $\Theta'_z$ in addition to $\Theta_d$ in eq~\eqref{eq:sentence_parameterize} as now the probability we are modelling is also conditioned on topics $\mathrm{z}_1, \dots, \mathrm{z}_K$:
\small
\begin{gather}
\mathrm{Pr}(\mathrm{y}|\mathrm{z}_1, \dots, \mathrm{z}_K, \mathrm{x}; \Theta_d, \Theta'_z) = \prod_{t=1}^{N_s} \mathrm{Pr}(w_t|w_{<t}, \mathrm{x}, \mathrm{z}_1, \dots, \mathrm{z}_K; \Theta_d, \Theta'_z) \nonumber \\ 
= \prod_{t=1}^{N_s} \psi_d(h_{t-1}, w_{t-1}, \mathrm{z}_1, \dots, \mathrm{z}_K; \Theta_d, \Theta'_z)
\end{gather}
\normalsize

Putting all these together, we get the parameterization for \emph{topic-aware conditional sentence probability} as:
\small
\begin{equation}
\label{eq:tgm_model}
\begin{aligned}
&\mathrm{Pr(y|x)} = \mathrm{Pr(y|x}; \Theta_d, \Theta'_z, \Theta_z) \\
&= \sum_{z_1, \dots, z_K}\psi_{z}(\mathrm{x}; \Theta_z) \prod_{t=1}^{N_s} \psi_d(h_{t-1}, w_{t-1}, \mathrm{z}_1, \dots, \mathrm{z}_K; \Theta_d, \Theta'_z)
\end{aligned}
\end{equation}
\normalsize
Contrasting with \emph{conditional sentence probability} parameterization in eq~\eqref{eq:sentence_parameterize}, we see that we have additional parameters $\Theta_z$ for topic prediction and $\Theta_d, \Theta'_z$ for topic-aware sentence modelling in the parameterization of \emph{topic-aware conditional sentence probability}. 

The standard loss function for caption generation task is to maximize the log probability of the \emph{conditional sentence probability}:
\small
\begin{equation}
\begin{aligned}
L_{caption}(\mathrm{y}, \tilde{\mathrm{y}}) &= - \sum_{t=1}^{N_s} \sum_{\tilde{w}_t \in \mathrm{W}}\delta_{w_t, \tilde{w}_t} \log \mathrm{Pr}(\tilde{w_t}|\mathrm{x}, w_{<t};\Theta_d)
\end{aligned}
\label{eqn:sentence_loss}
\end{equation}
\normalsize
where $\delta_{w_t, \tilde{w_t}}$ is the indicator function.
We could directly use this loss to train our topic-aware parameterized model in eq~\eqref{eq:topic_aware} as it is a special form of the general \emph{conditional sentence probability}. 
However in training, it requires large amount of data for the model to discover latent topics from scratch and to learn topic-aware sentence distribution simultaneously. 
The training data of caption task is limited,
and as the labelling cost of such training data is very high, the amount of the training data is not likely to grow very fast in the near future. 
To solve this problem, we introduce an auxiliary task, topic prediction, to guide the caption task. 
In this task, we leverage the existing topic mining approaches to generate a set of topics as the teacher topics $z$.
The details of the topic mining approach are given in Section~\ref{sec:topic_generation}.
We use the teacher topics to guide the latent topic learning by an additional topic loss function $L_{topic}(z, \tilde{z})$ on the topic predictor $\psi_t(\mathrm{x}; \Theta_z)$, which penalizes the latent topic predictions that are far away from the teacher $z$.
The details of $dist$ function in $L_{topic}(z, \tilde{z})$ will be discussed in section~\ref{sec:multitask}. 
The two losses are combined by the trading off hyper-parameter $\lambda \in [0, 1)$:
\small
\begin{gather}
L(\mathrm{y}, z, \mathrm{\tilde{y}}, \tilde{z}) = (1-\lambda) L_{caption}(\mathrm{y, \tilde{y}}) + \lambda L_{topic}(z, \tilde{z}) \label{eq:multitask_loss}\\
L_{caption}(\mathrm{y, \tilde{y}}) = - \sum_{t=1}^{N_s} \sum_{\tilde{w}_t \in \mathrm{W}} \delta_{w_t, \tilde{w}_t} \log \, \mathrm{Pr}(\tilde{w}_t|\mathrm{x}, w_{<t}; \Theta_d, \Theta'_z, \Theta_z) \label{eq:loss_sent}\\
L_{topic}(z, \tilde{z}) = dist(z, \tilde{z})  \label{eq:loss_topic}
\end{gather}
\normalsize
\begin{figure} \centering 
\includegraphics[width=\linewidth]{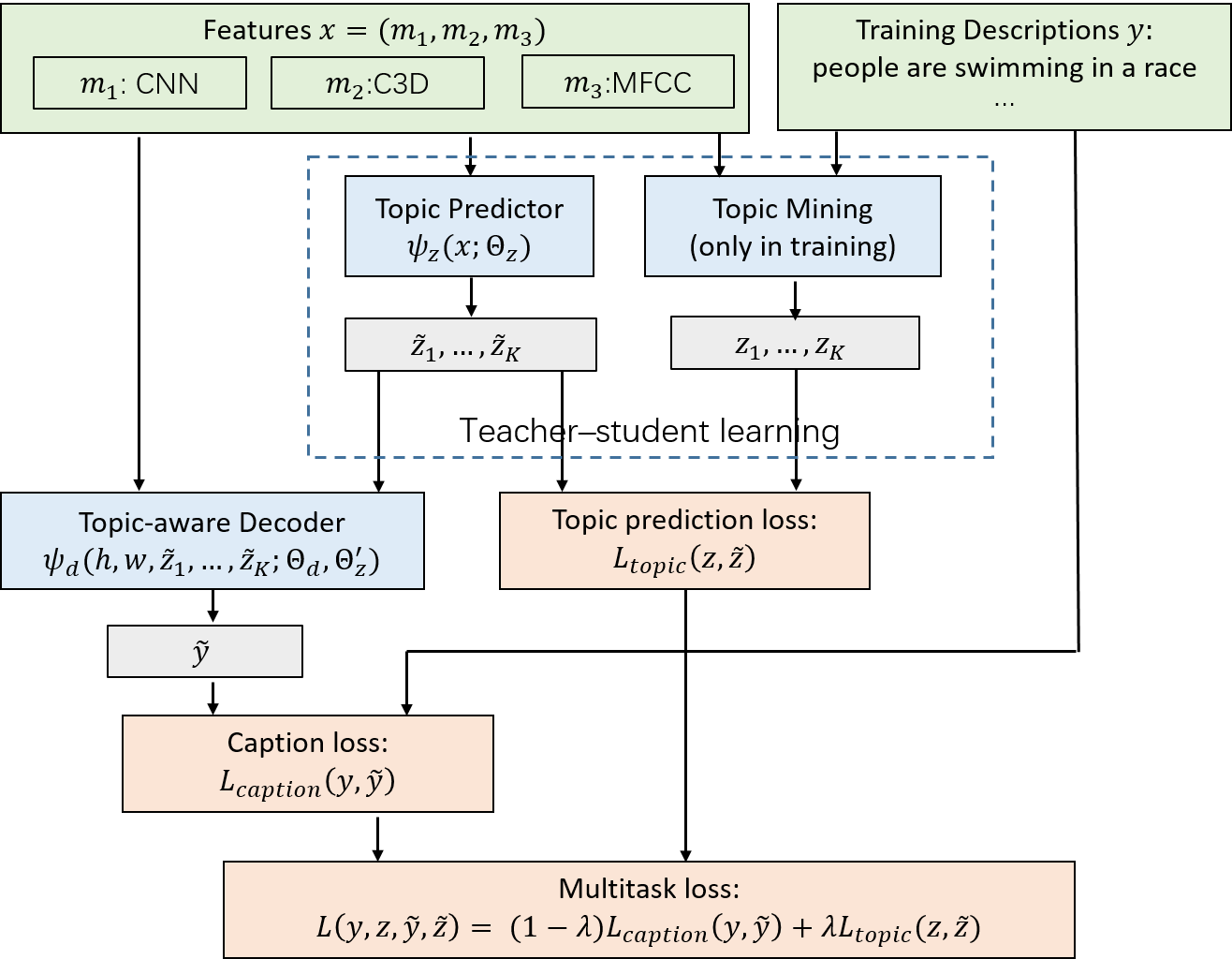} 
\caption{Framework of M\&M TGM. The green boxes are inputs; blue boxes are modules; gray boxes are outputs; red boxes are loss functions. The topic mining module is only used in the training stage while other modules are used in both training and testing. }
\label{fig:framework} 
\end{figure}
In this way, we weave the video latent topics into video description generation as guidance. 

\section{M\&M Topic-Guided Model}
In this section, we present our solutions for the topic-guided video captioning.
Figure~\ref{fig:framework} illustrates the overall framework.
Our proposed M\&M TGM consists of three components: topic mining, topic predictor and topic-aware decoder,
which is optimized by multi-task training.
We will first present the overall multi-task training scheme and then introduce the three modules in details.

\subsection{Multi-task Training Scheme}
\label{sec:multitask}

It is intractable to integrate the \emph{topic-aware conditional sentence probability} in eq~\eqref{eq:tgm_model} over the latent topic space.
But since the latent topic is highly dependent on the video content, usually the latent topic distributions of the video could be very skew with massive likelihood on the peak point.
Therefore, it is reasonable that we use the maximum latent topic likelihood to approximate the summation over the latent topic space to gain efficiency without losing much accuracy.
The approximation for eq~\eqref{eq:tgm_model} is as follows:
\begin{gather}
\mathrm{Pr}(\mathrm{y}|\mathrm{x}) = \sum_{\mathrm{z}_1, \dots, \mathrm{z}_K} \mathrm{Pr}(\mathrm{y}|\mathrm{z}_1, \dots, \mathrm{z}_K,\mathrm{x}) \mathrm{Pr}(\mathrm{z}_1, \dots, \mathrm{z}_K|\mathrm{x})\\
 \approx \mathrm{Pr}(\mathrm{y}|\mathrm{z}_1=\tilde{z}_1, \dots, \mathrm{z}_K = \tilde{z}_K, \mathrm{x}) \mathrm{Pr}(\mathrm{z}_1 = \tilde{z}_1, \dots, \mathrm{z}_K= \tilde{z}_K|\mathrm{x}) \nonumber
\label{eq:approximate_goal}
\end{gather}
where $\tilde{z_1}, \dots, \tilde{z_K} = \argmax_{\mathrm{z}_1, \dots, \mathrm{z}_K} \mathrm{Pr}(\mathrm{z}_1, \dots, \mathrm{z}_K|\mathrm{x})$. 
Our experiments show such approximation leads us to good results.

The multi-task training scheme is conducted as follows:

1. We first pretrain the topic predictor using the topic loss $L_{topic}$,
and use it to initialize $\Theta_z$ in the M\&M TGM (as shown in eq~\eqref{eq:tgm_model}).
For the $dist$ in topic loss, we choose two widely used distances: $l_{2}$-distance $l_{2}(z, \tilde{z}) = \parallel z - \tilde{z} \parallel_{2}^{2}$ and KL-divergence $KL(z, \tilde{z}) = \sum_{k=1}^{K} \tilde{z}_{k} log (\tilde{z}_{k}/{z_{k}})$.

2. We then pretrain the topic-aware decoder using the caption loss $L_{caption}$ (as shown in eq~\eqref{eq:loss_sent}) with the topic guidance from the fixed topic predictor in the above step, and use it as a good initialization for $\Theta'_{z}$ and $\Theta_d$ in the M\&M TGM (as shown in eq~\eqref{eq:tgm_model}).

3. Finally, based on the parameters initialization in the above two steps, 
we use the multi-task loss $L(y, z, \tilde{y}, \tilde{z})$ (as shown in eq~\eqref{eq:multitask_loss}) to further train both models jointly, which fine-tunes all the parameters $\Theta_z$, $\Theta'_{z}$ and $\Theta_d$ to optimize the caption performance.

\subsection{Topic Mining and Prediction}
\label{sec:topic_generation}
We first briefly present two topic structures adopted in previous works,
and then propose our multimodal latent topic mining approach and the topic prediction model.

\begin{figure}\centering
\subfigure[]{ \label{fig:category_tags}
\includegraphics[width=0.22\linewidth]{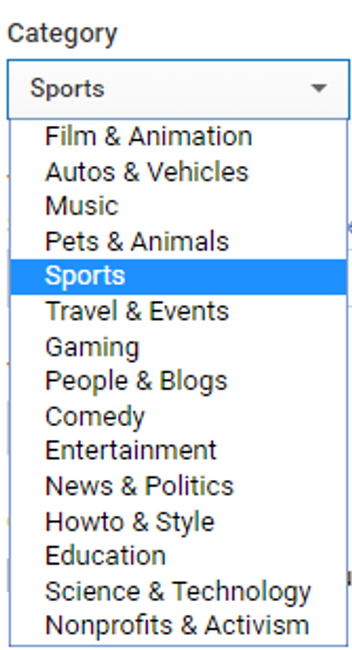}
}	
\subfigure[]{ \label{fig:text_topic_cons}
\includegraphics[width=0.5\linewidth]{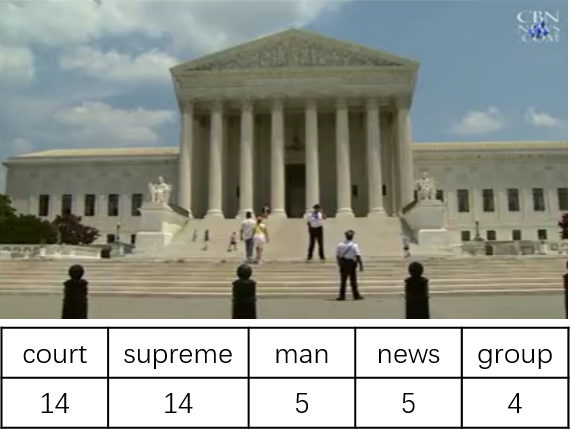}
}
\caption{(a) Expert-defined topic schema on the YouTube website. (b) An example video in MSR-VTT dataset with its top frequent caption words.} 
\end{figure}

\textbf{Expert-defined Topics:}
The online video websites often provide an expert-defined topic schema as shown in Figure~\ref{fig:category_tags}.
Video uploaders can select one of the topics to better organize their videos.
Such topics can reflect the topic variety to some extent, but have the following drawbacks:
1) The user assigned labels are noisy with labelling mistakes;
2) The topic distributions are exclusive which ignores the topic diversity inside the video;
and 3) There might exist different semantic meanings and visual appearances within a topic, which is suboptimal as latent topics.

\textbf{Textual Latent Topics:}
The annotated descriptions in video caption datasets provide rich and accurate information about the video content, which can also reflect more appropriate latent topic distributions.
Thus, our previous work \cite{DBLP:conf/mir/ChenCJ17} mined topics from the annotated video captions on the training set. 
The main idea is to cluster the video captions and each cluster can represent a latent topic.
We use the kernel K-means algorithm \cite{dhillon2004kernel} for clustering.
We group all the groundtruth captions of a video as one data sample.
Stopwords are removed and the bag-of-words are used as our text features.
The cosine kernel is adopted to generate nonlinear cluster separations.
Since a video might contain several topics, we utilize the soft assignment according to the distance between samples and clusters to generate topic distributions.

\textbf{Multimodal Latent Topics:}
Although the textual latent topics may better fit with the videos' underlying topic distributions, the mined topics purely based on textual data still suffer from the following problems:

1) Polysemy phenomena: Words can convey several different meanings, which might lead to improper assignment of the video.
As shown in Figure~\ref{fig:text_topic_cons}, the most frequent word \textit{court} can either mean tribunal or sports field, so the video could mistakenly peak at some sports related topics.

2) Certain messy clusters: The unsupervised clustering method cannot perfectly generate meaningful latent topics, which could produce some semantically unclear topics with dissimilar visual appearances. This brings great harm for the topic prediction model to learn from those messy topics.

To address these issues, we further propose to combine the description texts and visual features to learn the multimodal latent topic representation.
The textual features are preprocessed in the same way as in textual latent topics and the visual features are elaborated in Section~\ref{sec:implementation_details}.
We use the weighted kernel K-means algorithm to fuse the textual and visual features with weights of 1 and 0.2 respectively, because we consider the textual features are more accurate than the visual features to unveil an eligible latent topic structure.

\textbf{Topic Predictor:}
We take the teacher-student perspective \cite{Ba2013Do} to train the topic predictor $\psi_{z}$.
The distributions of the above mined latent topics serve as the teacher $z$ to guide the $\psi_{z}$ to generate latent topic predictions with the topic prediction loss $L_{topic}$.
In this work, we adopt a two-layer perception as the topic predictor based on inputs of multimodal video features $m_1, m_2, m_3$:
\begin{equation}
\psi_z(\mathrm{x}; \Theta_z) = softmax(W_{2}(\varphi(W_{1}[m_{1};m_{2};m_{3}]+b_{1}) + b_{2})
\end{equation}
where $W_{i}, b_{i} (i=1,2)$ are parameters, $[\cdot]$ denotes feature concatenation and $\varphi$ is the RELU nonlinear function \cite{glorot2011deep}.


\subsection{Topic-aware Decoder}
\label{sec:topic_decoder}

In this section, we first briefly introduce the standard LSTM, and then describe our proposed topic-aware decoder based on the LSTM model to employ the topic guidance during sentence generation.

The LSTM model addresses the vanishing gradients problem in traditional RNN by employing a memory cell and three gates to control the information flow in the network.
The formulas of the LSTM cell $\psi_d(h_t, w_t; \Theta_d)$ at timestep $t$ are given below:
\begin{align}
input\ gate:\ &i_{t} = \sigma(W_{iw}w_{t}+U_{ih}h^{hidden}_{t-1}+b_{i}) \\
forget\ gate:\ &f_{t} = \sigma(W_{fw}w_{t}+U_{fh}h^{hidden}_{t-1}+b_{f}) \\
output\ gate:\ &o_{t} = \sigma(W_{ow}w_{t}+U_{oh}h^{hidden}_{t-1}+b_{o}) \\
cell\ input:\ &g_{t} = \phi(W_{gw}w_{t}+U_{gh}h^{hidden}_{t-1}+b_{g}) \\
cell\ state:\ &h^{cell}_{t} = i_{t} \odot g_{t} + f_{t} \odot h^{cell}_{t-1} \\
cell\ output:\ &h^{hidden}_{t} = o_{t} \odot \phi(h^{cell}_{t})
\end{align}
where $h^{hidden}_t$ is the hidden state, $h^{cell}_t$ is the cell state, $\sigma$ is sigmoid function, $\phi$ is $\tanh$ function, $\odot$ denotes element-wise production, and $\Theta_d = \{W_{*w}, U_{*h}, b_{*}\}$ are parameters.

Recall that in Section~\ref{sec:problem_form}, when we use LSTM cell to parameterize $\mathrm{Pr(y|z_1, \dots, z}_K, \mathrm{x})$, we need to change the LSTM cell function to $\psi_d(\mathrm{w}_t, h_t, \mathrm{z}_1, \dots, \mathrm{z}_K; \Theta_d, \Theta'_z)$ to employ the topic dependency. 
Inspired by Gan et al. \cite{Gan2016Semantic}, we extend each weight matrix of the conventional LSTM to be an ensemble of a set of topic-dependent weight matrices.

Let us take one of the input gate weight matrices $W_{iw}$ as an example, and transformation for other parameters in LSTM model are alike.
We define the ensemble 3D weight matrix $\mathrm{W}_{i\tau} \in \mathbb{R}^{n_{h}\times n_{w} \times K}$, where $n_{h}$ is the number of hidden units and $n_{w}$ is the dimension of input vectors.
$\mathrm{W}_{i\tau}[k]$ denotes the $k$-th slice of $\mathrm{W}_{i\tau}$, which represents the 2D weight matrix belonging to the LSTM model for the $k$-th topic.
Therefore, we explicitly specify $K$ language decoders, which is, for each topic there is a pair of topic-specific LSTM weights.
Given the latent topic $\tilde{z}$, we can define the mixture topic LSTM weight matrix $\mathrm{W}_{i}(\tilde{z}) \in \mathbb{R}^{n_{h}\times n_{w}}$ as 
\begin{equation}
\mathrm{W}_{i}(\tilde{z}) = \sum_{k=1}^{K} \tilde{z}_{k}\mathrm{W}_{i\tau}[k]
\end{equation}
where $\tilde{z}_{k}$ is the $k$-th topic in $\tilde{z}$.
When $\tilde{z}$ is not the exclusive one-hot topic distribution, the video data is shared among different topics.
However, the parameters still grow linearly with the number of topics $K$ and no parameters are shared among different video topics which can easily result in over-fitting. 
So the 3-way factorization method \cite{Memisevic2007Unsupervised,krizhevsky2010factored} is used to share parameters.
We re-represent $\mathrm{W}_{i\tau}$ in terms of three matrices $\mathrm{W}_{ia} \in \mathbb{R}^{n_{h}\times n_{f}}$, $\mathrm{W}_{ib} \in \mathbb{R}^{n_{f}\times K}$ and $\mathrm{W}_{ic} \in \mathbb{R}^{n_{f}\times n_{w}}$, where $n_{f}$ is the number of factors, such that
\begin{equation}
\mathrm{W}_{i}(z) = \mathrm{W}_{ia} \cdot diag(\mathrm{W}_{ib}\tilde{z}) \cdot \mathrm{W}_{ic}
\end{equation} 
$\mathrm{W}_{ia}$ and $\mathrm{W}_{ic}$ are shared among all topics, while $\mathrm{W}_{ib}$ can be viewed as the latent vectors of topics.

Therefore, the transformation from input vectors to input gates is changed from $\mathrm{W}_{iw} \cdot w_{t}$ to 
$\mathrm{W}_{ia} \cdot (\mathrm{W}_{ib}\tilde{z} \odot \mathrm{W}_{ic}w_{t})$.
The topic distribution $\tilde{z}$ affects the LSTM parameters associated to the video when decoding, which implicitly works as an ensemble of $K$ topic-aware language decoders.

\section{Experiments}
\begin{table*}
\centering
\caption{Caption performance comparison between Vanilla (no topic guidance) and TGM (using different latent topic guidance) on the MSR-VTT and Youtube2Text datasets. The best results are marked in bold and the second best results with \underline{underline}.}
\renewcommand{\arraystretch}{1.1}
\begin{tabular}{|c|c|cccc|cccc|} \hline
\multirow{2}{*}{model} & \multirow{2}{*}{topic guidance} & \multicolumn{4}{c|}{MSR-VTT} & \multicolumn{4}{c|}{Youtube2Text} \\ \cline{3-10}
 &  & \small BLEU4 & \small METEOR & \small ROUGE$_{l}$ & \small CIDEr & \small BLEU4 & \small METEOR & \small ROUGE$_{l}$ & \small CIDEr \\ \hline
Vanilla & no topic & 42.23 & 28.68 & 61.44 & 46.06 & 46.05 & 32.97 & 69.94 & 72.12 \\ \hline
\multirow{4}{*}{TGM} 
 & pred expert-defined topic & 42.36 & 28.51 & 61.57 & 46.37 & 47.42 & 33.51 & 69.61 & 77.36 \\
 & pred textual latent topic & \underline{43.38} & \underline{29.08} & \underline{62.11} & 48.38 & \underline{47.45} & \underline{34.11} & \textbf{70.59} & \underline{79.45} \\
 & pred multimodal latent topic & \textbf{43.81} & \textbf{29.26} & \textbf{62.13} & \underline{48.50} & \textbf{47.56} & \textbf{34.21} & \underline{70.45} & \textbf{79.57} \\ \cline{2-10}
 & assigned expert-defined topic & 42.75 & 29.07 & 61.77 & \textbf{48.59} & 46.47 & 32.99 & 69.20 & 74.24 \\ \hline
\end{tabular}
\label{tab:topics_comparison}
\end{table*}

\subsection{Datasets}
To validate the effectiveness, robustness and generalization of our proposed methods, we conduct extensive experiments on two benchmark video captioning datasets: MSR-VTT \cite{xu2016msr} and Youtube2Text \cite{DBLP:conf/iccv/GuadarramaKMVMDS13}.

\begin{figure} \centering 
\includegraphics[width=1\linewidth]{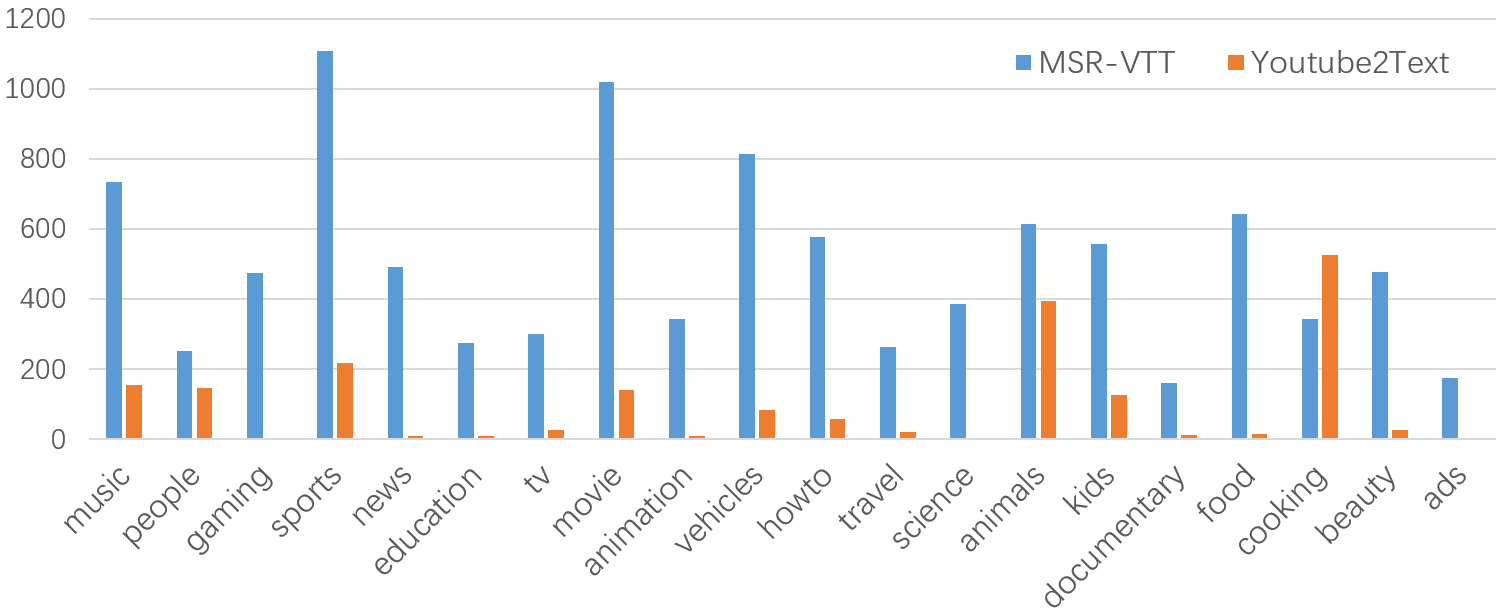} 
\caption{The distribution of noisy expert-defined topics in MSR-VTT and Youtube2Text datasets.} 
\label{fig:dataset_categories} 
\end{figure}

\textbf{MSR-VTT}: The MSR-VTT corpus \cite{xu2016msr} is currently the largest open-domain video captioning dataset with a wide variety of video topics.
It consists of 10,000 video clips with 20 human annotated captions per clip.
Each clip also contains a noisy expert-defined topic label crawled from YouTube, the distribution of which is shown in Figure~\ref{fig:dataset_categories}.
Following the standard data split in the MM2016 challenge \cite{VTTGC2016}, we use 6,513 videos for training, 497 videos for validation and the remained 2,990 for testing.

\textbf{Youtube2Text}: The Youtube2Text corpus \cite{DBLP:conf/iccv/GuadarramaKMVMDS13} contains 1970 Youtube video clips with around 40 human annotated sentences per clip.
For fair comparison, we crawl the expert-defined topics from YouTube for the whole dataset as shown in Figure~\ref{fig:dataset_categories}, which are also noisy.
We adopt the same data splits as provided in Yao et al. \cite{DBLP:conf/iccv/YaoTCBPLC15}, with 1,200 videos for training, 100 videos for validation and 670 videos for testing. 

\textbf{Description Preprocessing}: We convert all descriptions to lower case and remove all punctuations.
We add begin-of-sentence tag $<$BOS$>$ and end-of-sentence tag $<$EOS$>$ to our vocabulary.
We preserve words that appear more than twice for MSR-VTT, resulting in a vocabulary size of 10,868,
and words that appear more than once for Youtube2Text, resulting in a vocabulary size of 7,245.
\subsection{Implementation Details}
\label{sec:implementation_details}

\textbf{Multi-modality Features}:
We extract features from image, motion and aural modalities.
For image features, we extract activations from the penultimate layer of the inception-resnet \cite{Szegedy2016Inception} pre-trained on the ImageNet, the dimensionality of which is 1,536. 
For motion features, we extract activations from the last 3D convolution layer of the C3D model \cite{tran2015learning} pre-trained on the Sports-1M dataset.
We perform max-pooling on the spatial dimension (width and height), resulting in 512 dimensional features. 
For aural features, We extract the Mel-Frequency Cepstral Coefficients (MFCCs) \cite{davis1980comparison} and use Bag-of-Audio-Words \cite{pancoast2014softening} and Fisher Vector \cite{sanchez2013image} encoding methods to generate video-level features, with dimensionality of 1,024 and 624 respectively. 
We simply pad zeros for videos without the sound track.

\textbf{Training Settings}:
We empirically set the hidden layer of the topic prediction model with 512 units.
The dimension of the LSTM hidden size is set to be 512.
The output weights to predict the words are the transpose of the input word embedding matrix.
We apply dropout with rate of 0.5 on the input and output of LSTM and use ADAM algorithm \cite{kingma2014adam} with learning rate of $10^{-4}$.
Beam search with beam width of 5 is used to generate sentences during testing process.

\textbf{Evaluation Metrics}:
We evaluate the caption results comprehensively on all major metrics, including BLEU \cite{papineni2002bleu}, METEOR \cite{denkowski2014meteor}, ROUGE-L \cite{lin2004rouge} and CIDEr \cite{vedantam2015cider}.

\subsection{Evaluation of M\&M TGM}
In this subsection, we introduce two baselines to show the effectiveness of topic guidance and our multi-task learning for video captioning. 
The first baseline, \emph{Vanilla}, consists of the multimodal encoder and the standard LSTM decoder. This baseline doesn't involve any topic information. 
The second baseline is \emph{TGM}, a trimmed version of our M\&M TGM. We discard the multi-task joint optimization step (last step) in section~\ref{sec:multitask}. 

To evaluate the effectiveness of topic guidance, we compare the caption performance of the Vanilla and TGMs using different latent topic guidance in Table~\ref{tab:topics_comparison}.
The numbers of textual and multimodal latent topics are optimized on the validation set (50 topics for MSR-VTT and 5 topics for Youtube2Text).
We can see that TGMs outperform Vanilla model consistently on all four metrics across MSR-VTT and Youtube2Text datasets, which demonstrates that exploiting topic information is beneficial to generate video descriptions.

The first three rows in the TGM block in Table~\ref{tab:topics_comparison} shows a fair comparison of the impact from different predicted latent topics on caption performance. 
We can see that the automatically mined topics (both textual and multimodal latent topics), outperform expert-defined topics by a large margin on all four metrics and two datasets, such as over 2\% absolute gain on the CIDEr score.
The multimodal latent topics further achieve better performance than textual latent topics consistently on multiple metrics across datasets. 
The improvement is also proved to be significant in the Student's t-test such as p-value of 0.03 on the BLEU$@$4 score.
These results suggest that the multimodal topic mining approach can discover better underlying topics as guidance for TGM,
and such improvement is prominent and robust for different evaluation metrics and datasets.

\begin{table} 
\centering
\renewcommand{\arraystretch}{1.1}
\caption{Caption performance of TGM (single-task learning) and M\&M TGM (multi-task learning) on MSR-VTT and Youtube2Text datasets.}
\label{tab:multitask_expr}
\begin{tabular}{|c|c|ccc|} \hline
dataset & model &  BLEU4 &  METEOR &  CIDEr \\ \hline
\multirow{2}{1.1cm}{MSR-VTT} & TGM & 43.81 & 29.26 & 48.50 \\
& M\&M TGM & \textbf{44.33} & \textbf{29.37} & \textbf{49.26} \\ \hline
\multirow{2}{1.1cm}{Youtube 2Text} & TGM & 47.56 & 34.21 & 79.57 \\
& M\&M TGM & \textbf{48.76} & \textbf{34.36} & \textbf{80.45} \\ \hline
\end{tabular}
\end{table}

Since video uploaders can manually assign expert-defined topic labels on online websites, we also compare our automatically predicted multimodal latent topics with the user assigned expert-defined topics in the last two rows of Table~\ref{tab:topics_comparison}.
However, the assigned expert-defined topics are very noisy, because the annotations are labelled on the entire videos but the caption generation is only applied on the short segments and the users may also make labelling mistakes.
For almost all metrics, the performances of our predicted latent topics are superior to the noisy assigned expert-defined topics on both MSR-VTT and Youtube2Text datasets, 
which shows that the latent topics learned in an unsupervised approach are competitive or even better than the assigned expert-defined topics with noise.

To validate the influence of joint training strategy, we compare the caption performance of TGM and M\&M TGM.
As shown in Table~\ref{tab:multitask_expr},
M\&M TGM consistently improves TGM on all datasets and metrics, with significance p-value $< 0.05$ for different metrics in Student's t-test,
which proves the benefits brought by the multi-task learning.


\subsection{Comparison with the State-of-the-art}
\label{sec:state-of-art-cmpr}
\begin{table} \small
\centering
\renewcommand{\arraystretch}{1.1}
\caption{Caption performance of M\&M TGM and state-of-the-art methods on MSR-VTT and Youtube2Text datasets.}
\label{tab:state_of_art_cmpr}
\begin{tabular}{|c|c|ccc|} \hline
dataset & model & BLEU4 & METEOR & CIDEr \\ \hline
\multirow{4}{0.9cm}{MSR-VTT} 
& M\&M TGM & \textbf{44.33} & \textbf{29.37} & \textbf{49.26} \\ \cline{2-5}
& Aalto \cite{shetty2016frame} & 39.80 & 26.90 & 45.70 \\
& v2t\_navigator\footnotemark[1] \cite{DBLP:conf/mm/JinCCXH16} & 40.80 & 28.20 & 44.80 \\
& dense caption \cite{cvpr17densevideocaption} & 41.40 & 28.30 & 48.90 \\ \hline \hline
\multirow{9}{0.9cm}{Youtube 2Text} 
& M\&M TGM & 48.76 & \textbf{34.36} & \textbf{80.45} \\ \cline{2-5}
& LSTM-YT \cite{Venugopalan2014Translating} & 33.30 & 29.10 & - \\
& S2VT \cite{venugopalan2015sequence} & - & 29.80 & - \\
& LSTM-I \cite{dong2017improving} & 44.60 & 31.10 & - \\
& SA \cite{DBLP:conf/iccv/YaoTCBPLC15} & 41.92 & 29.60 & 51.67 \\
& LSTM-E \cite{DBLP:conf/cvpr/PanMYLR16} & 45.30 & 31.00 & - \\ 
& h-RNN decoder \cite{DBLP:conf/cvpr/YuWHYX16} & 49.90 & 32.60 & 65.80 \\
& h-RNN encoder \cite{pan2015hierarchical} & 43.80 & 33.10 & - \\ 
& SCN-LSTM \cite{Gan2016Semantic} & \textbf{50.20} & 33.40 & 77.70 \\ \hline
\end{tabular}
\end{table}
\footnotetext[1]{winner of the MM16 VTT challenge.}

Table~\ref{tab:state_of_art_cmpr} presents our M\&M TGM with several state-of-the-art methods applied on the two video caption datasets.
Our method achieves significant better performance than prior works on MSR-VTT dataset, for example, the BLEU$@$4 achieves 7.08\% relative improvement than the previous best performance.
For Youtube2Text dataset, the METEOR and CIDERr scores improved significantly but our performance on BLEU$@$4 is lower than the SCN-LSTM.
The BLEU$@$4 metric focuses on the syntactic agreement, while METEOR and CIDEr concern more about semantic meanings.
Therefore, we consider the sentence generated from M\&M TGM are more semantically relevant with the video content.

To be noted, the semantic concepts used in SCN-LSTM are trained with additional image data because they claim Youtube2Text corpus is too small to train reliable concept classifiers,
but our M\&M TGM is purely trained on Youtube2Text for latent topic prediction and sentence generation.
Hence, we argue that for video captioning, the guidance from latent topics might be superior than detected semantic concepts for the following reasons:
1) videos contain more objects than images but many might be irrelevant to the description;
2) topics contain additional information besides concepts such as actions from motion modality and words from speech modality;
and 3) the prediction accuracy is important for the decoder and topics are easier to be classified than concepts.

\begin{table}
\centering
\renewcommand{\arraystretch}{1.1}
\caption{Comparison of generalization ability of the vanilla model and our proposed M\&M TGM model. The models are trained on MSR-VTT training set and evaluated on the Youtube2Text testing set.}
\label{tab:vtt_msvd_generalization}
\begin{tabular}{|c|cccc|} \hline
 & BLEU4 & METEOR & ROUGE$_{l}$ & CIDEr \\ \hline
Vanilla & 32.92 & 30.16 & 62.51 & 51.02 \\
M\&M TGM & \textbf{34.67} & \textbf{30.68} & \textbf{63.68} & \textbf{55.39} \\ \hline
\end{tabular}
\end{table}

\begin{table}
\centering
\renewcommand{\arraystretch}{1.1}
\caption{Caption performance comparison of different topic prediction losses.}
\label{tab:topic_loss_comparison}
\begin{tabular}{|c|c|cccc|} \hline
dataset & loss & BLEU4 & METEOR & ROUGE$_{l}$ & CIDEr \\ \hline
\multirow{2}{1.1cm}{MSR-VTT} & $l_{2}$ & \textbf{43.81} & \textbf{29.26} & \textbf{62.13} & 48.50 \\
& KL & 43.40 & 29.11 & 62.01 & \textbf{48.64} \\ \hline
\multirow{2}{1.1cm}{Youtube 2Text} & $l_{2}$ & \textbf{47.56} & \textbf{34.21} & \textbf{70.45} & \textbf{79.57} \\
& KL & 45.98 & 33.65 & 69.76 & 78.06 \\ \hline
\end{tabular}
\end{table}

\subsection{Experimental Analysis}
\label{subsec:expr_analysis}
\textbf{Generalization Ability:}
To evaluate the generalization ability of our proposed method, we conduct the cross dataset experiment.
We train M\&M TGM on MSR-VTT dataset and test its performance on the Youtube2Text dataset.
Results are presented in Table~\ref{tab:vtt_msvd_generalization}.
We can see that the proposed M\&M TGM works significantly better than Vanilla model for cross datasets evaluation on all four metrics, which demonstrates that our method not only can improve the caption performance for in-domain videos but also generalize well on videos in the wild.

\textbf{Topic Loss Selection:}
We compare the caption performance using $l_{2}$-distance or KL-divergence as $L_{topic}$ in the TGM model.
Table~\ref{tab:topic_loss_comparison} presents the results.
The $l_{2}$-distance consistently surpasses the KL-divergence across datasets and metrics.
So unless otherwise specified, we use $l_{2}$-distance as the topic prediction loss.

\begin{figure}\centering
\subfigure[MSR-VTT]{ \label{fig:vtt_num_topics_cider}
\includegraphics[width=0.45\linewidth]{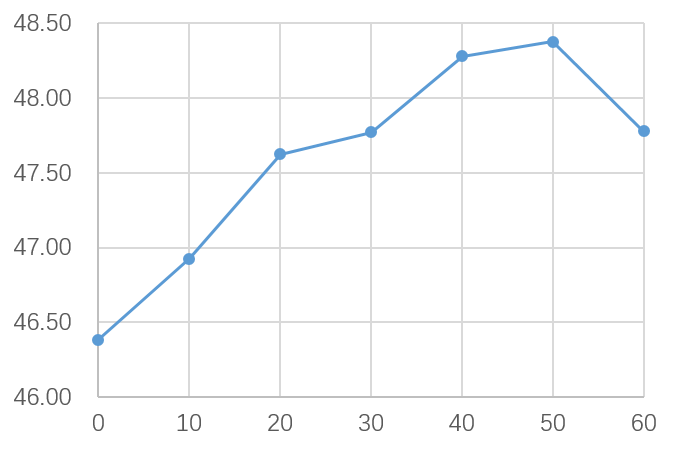}
}	
\subfigure[Youtube2Text]{ \label{fig:msvd_num_topics_cider}
\includegraphics[width=0.42\linewidth]{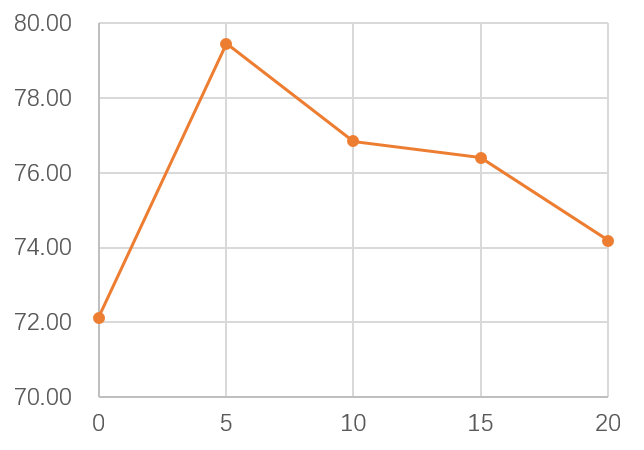}
}
\caption{CIDEr scores with number of topics on (a) MSR-VTT and (b) Youtube2Text datasets.} 
\label{fig:num_topics_cider}
\end{figure}

\textbf{The Number of Topics $K$:}
We also explore different number of topics on each dataset and the results are shown in Figure~\ref{fig:num_topics_cider}.
The best number of topics for MSR-VTT is 50 and for Youtube2Text is 5, which shows that the more diverse the dataset is, the more amount of topics is required.

\begin{figure}\centering
\subfigure[MSR-VTT]{ \label{fig:vtt_multitask_weight}
\includegraphics[width=0.42\linewidth]{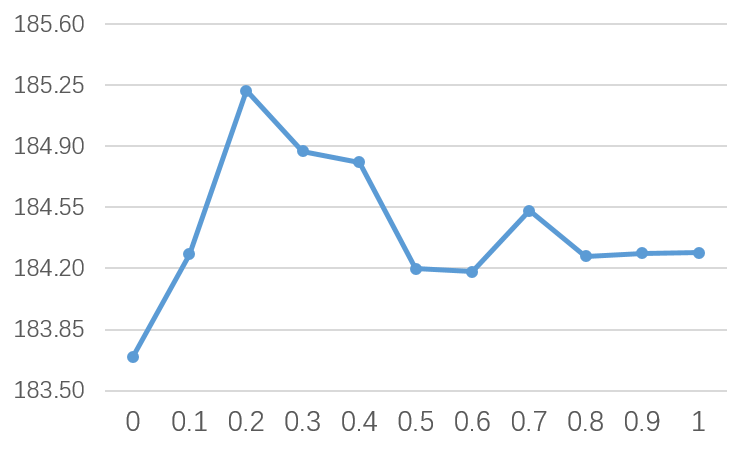}
}	
\subfigure[Youtube2Text]{ \label{fig:msvd_multitask_weight}
\includegraphics[width=0.42\linewidth]{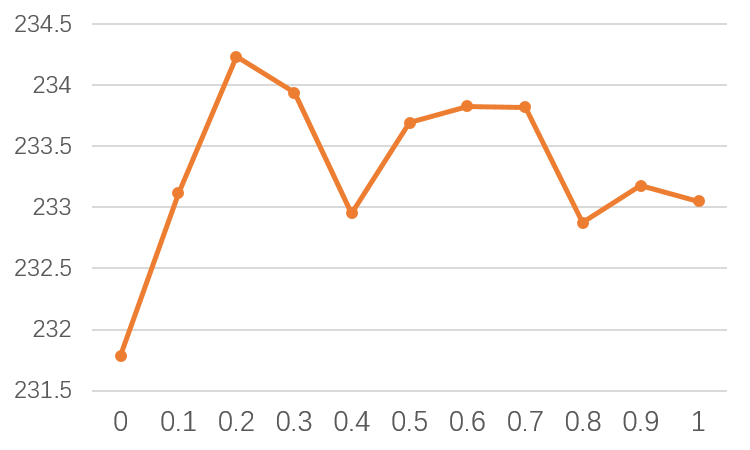}
}
\caption{The caption performance (sum of all metrics) with different multi-task hyper-parameter $\lambda$.} 
\label{fig:multitask_weight}
\end{figure}

\textbf{The Multi-task Parameter $\lambda$:}
Since caption generation is our main goal, we consider that the weight on topic prediction loss should not surpass the weight on caption generation loss.
Therefore, we search the best multi-task hyper parameter $\frac{\lambda}{1-\lambda}$ in range $[0.1, 1]$ with step of 0.1.
To save space, we present the sum of all the metrics in Figure~\ref{fig:multitask_weight} and the trends are similar for different metrics.
The multi-task learning with different $\lambda>0$ all achieves better performance than single-task learning ($\lambda=0$), which proves the robustness of our multi-task training with respect to the hyper-parameter.

\subsection{Qualitative Analysis}
To gain an intuition of the improvement on generated video descriptions from M\&M TGM model,
we present some video examples with the video description from Vanilla model and M\&M TGM on testing set of MSR-VTT.

\begin{figure}\centering
\includegraphics[width=0.93\linewidth]{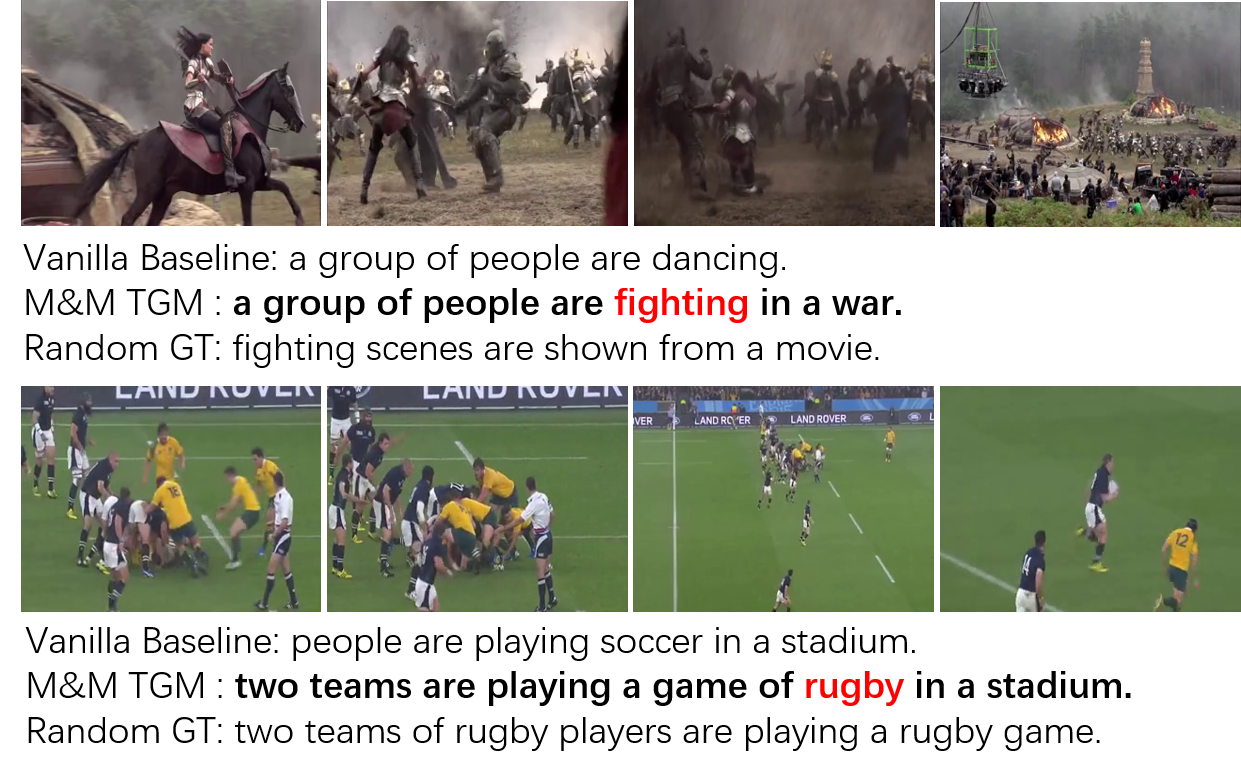}
\caption{Examples on MSR-VTT testing set. M\&M TGM can generate more accurate descriptions than Vanilla model.}
\label{fig:accurate_captions}
\end{figure}

\begin{figure}\centering
\includegraphics[width=0.93\linewidth]{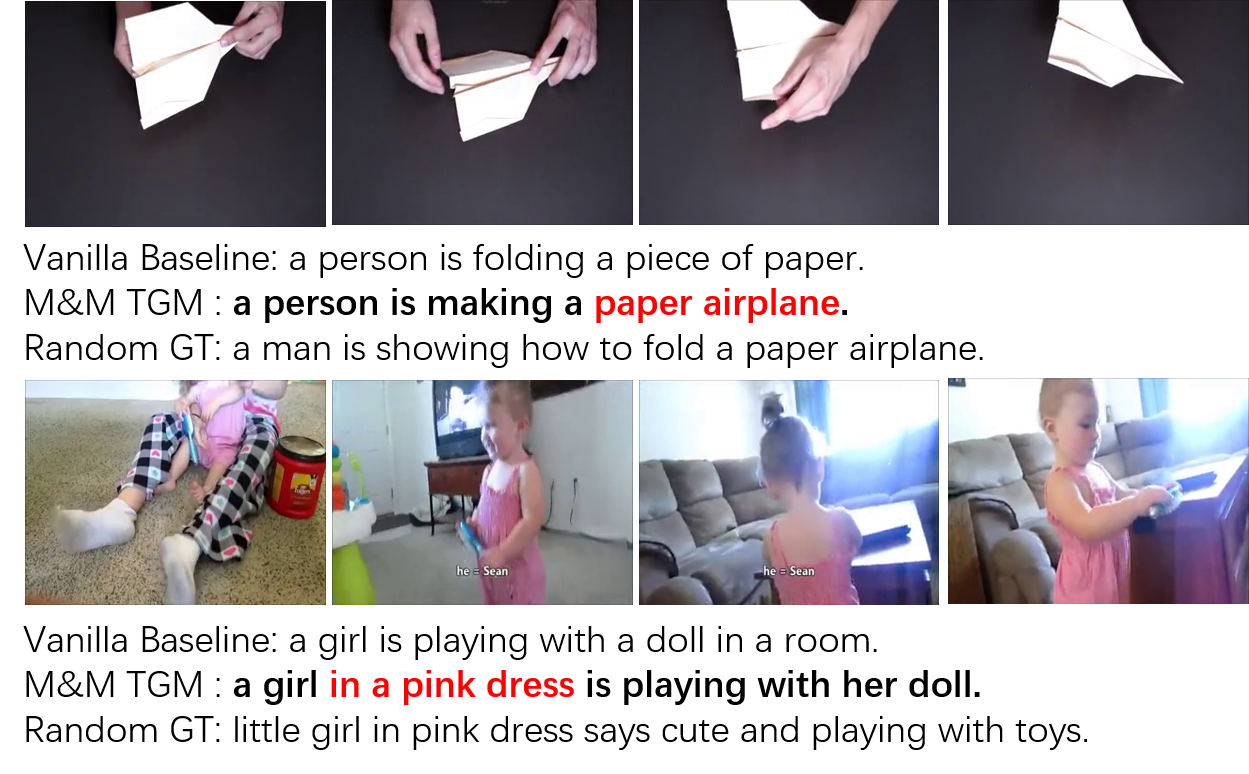}
\caption{Examples on MSR-VTT testing set. M\&M TGM can generate more detailed descriptions than Vanilla model.}
\label{fig:detailed_captions}
\end{figure}

In Figure~\ref{fig:accurate_captions}, we can see that M\&M TGM can generate more accurate video descriptions than Vanilla model even though they utilize the same multimodal features.
In Figure~\ref{fig:detailed_captions}, though the descriptions from Vanilla and M\&M TGM are both correct, M\&M TGM model can produce more detailed information about the video contents.
We also observe that M\&M TGM employs more unique words with 493 compared to the 391 unique words in Vanilla model.

The reason behind these quality improvements could be that M\&M TGM can narrow down the sentence generation space according to the predicted latent topics.
This enables the model to focus on subtler differences between similar concepts such as the soccer and rugby sports in the upper row of Figure~\ref{fig:accurate_captions},
and cover more detailed descriptions under the topic such as making airplane vs. folding paper in the upper row of Figure~\ref{fig:detailed_captions} with more specialized words.


\subsection{Human Interaction in Captioning}
Besides the improvement on the caption performance, our topic-guided model can provide an interface for users to be involved in the caption generation.
For example, users could simply assign a category tag to the uploaded videos with minimum costs to refine the automatic generated video descriptions.
To evaluate the performance boosts with the human enhancement to our topic-guided model, we manually re-annotate the expert-defined topics for Youtub2Text dataset as the clean version and compare its caption performance with the noisy assigned topics.
Results are shown in Table~\ref{tab:human_interaction} which demonstrates that a huge gain can be achieved with the manual correction of the topics in low cost.

\begin{table}
\caption{The influence of human interactions on Youtube2Text dataset with noisy and clean expert-defined topics assignment.}
\label{tab:human_interaction}
\begin{tabular}{|c|cccc|} \hline
 & \small BLEU4 & \small METEOR & \small ROUGE$_{l}$ & \small CIDEr \\ \hline
 noisy topic & 46.47 & 32.99 & 69.20 & 74.24 \\
clean topic & \textbf{49.35} & \textbf{35.10} & \textbf{71.14} & \textbf{82.79} \\  \hline
\end{tabular}
\end{table}
\section{Conclusions}

In this paper, we propose a novel topic-guided caption model to address the topic diversity challenge for open-domain video captioning task.
The proposed model can predict the latent topics of videos and then generate topic-oriented video descriptions with the topic guidance jointly in an end-to-end manner.
We utilize the multimodal topic mining approach to construct video topics automatically and take a teacher-student learning perspective to predict the latent topics purely from video multimodal contents.
The topic-aware decoder can exploit the predicted topics to adjust its weights to fit the topic-dependent sentence distributions.
Our experimental results on two public video caption benchmark datasets show that the proposed model can generate more accurate and detailed descriptions within different topics and improves the performance consistently on all metrics on both datasets. 
Furthermore, we show that our model has very good generalization ability across datasets. 
The proposed topic-guided caption model can be considered as a generic framework, which could be integrated with other techniques such as temporal attention or hierarchical video encoder. We will study such integration in the future. 

\section{Acknowledgments}
This work is supported by National Key Research and Development Plan under Grant No. 2016YFB1001202.

\bibliographystyle{ACM-Reference-Format}
\bibliography{reference} 

\end{document}